\documentclass[runningheads]{llncs}

\usepackage[T1]{fontenc}
\usepackage{graphicx}
\graphicspath{{../}}
\usepackage{booktabs}
\usepackage{amsmath}
\usepackage{amssymb}
\usepackage{multirow}
\usepackage{xcolor}
\usepackage{hyperref}
\usepackage{float}
\usepackage{subfigure} 
\usepackage{subcaption} 
\usepackage{cite}

\newcommand{\seniorauthor}{\textsuperscript{*}}
\newcommand{\equalcontrib}{\textsuperscript{\#}}
\begin{document}

\title{ChatT2: An Adaptive Framework for Developing a Large Language Model-Based Agent for Natural Product Domain Research}

\author{
Yihan Wang\equalcontrib
\and
Qiandi Gao\equalcontrib
\and
Yihui Zhuang
\and
Liangjun Ge
\and
Heqian Zhang\seniorauthor
\and
Jiaquan Huang\seniorauthor
\and
Zhiwei Qin\seniorauthor
}

\authorrunning{Yihan Wang et al.}
\titlerunning{Yihan Wang et al.}

\institute{
Center for Biological Science and Technology, Advanced Institute of Natural Sciences, Beijing Normal University, Zhuhai, Guangdong, 519087, China. \\
\equalcontrib These authors contributed equally\\
\seniorauthor Correspondence: \texttt{z.qin@bnu.edu.cn}; \texttt{jiaquan\_terry@bnu.edu.cn}; \texttt{zhangheqian@bnu.edu.cn}
}

\maketitle

\begin{abstract}
Scientific investigations into microbial natural products (NPs) present significant challenges for novices, largely due to the complexity of microbial systems, biochemical diversity, technical skill requirements, and the demands of bioinformatics and data analysis processes. To address these issues, we introduce ChatT2, a large language model (LLM)-based agent that is specifically tailored to the unique characteristics of bacterial type II polyketides. These polyketides form a structurally distinct and therapeutically important NP family. ChatT2 was developed within an autonomous multiagent framework composed of a mentor, an executor, and an evaluator, each with defined responsibilities. The mentor acts as an intermediary between ChatT2 and the user, utilizing chain-of-thought prompting to refine the intent of the user. Under the guidance of the mentor, the executor synthesizes multimodal information via retrieval-augmented generation techniques and seamlessly integrates bioinformatics and cheminformatics tools. The evaluator ultimately assesses the output of the executor to ensure the richness and accuracy of the retrieved information. Our research highlights how ChatT2, designed with this multiagent framework, addresses the challenges faced by general LLMs in terms of understanding limited, specialized corpora and complex biological information and provides both experts and novices with a valuable tool for exploring various NPs of interest. The ChatT2 webserver can be accessed at \url{https://chatt2.site/#/chat}.

    
\end{abstract}

\section{Introduction}

Microbial natural products have received constant attention from the scientific community because of their relevance to drug discovery, chemical ecology and molecular biology in general. These products are classified into categories such as polyketides, nonribosomal peptides, ribosomally synthesized and post-translationally modified peptides, terpenoids, and alkaloids. Bacterial aromatic polyketides (type II polyketides, T2PKs) represent a significant class of bioactive compounds that are renowned for their structural diversity and potent pharmacological properties. These metabolites, which are synthesized primarily by type II polyketide synthases (PKSs), have contributed to the development of antibiotics and antitumour drugs. In the field of natural product chemistry, the abundance of microbial-derived natural product data alongside associated article texts, such as the Natural Products Atlas, Collection of Open Natural Products database, and Minimum Information about a Biosynthetic Gene cluster (MIBiG), offers a rich landscape for mining data and extracting information~\cite{cech2021benefiting}. The development of comprehensive natural product databases has significantly increased the productivity of researchers by harnessing the wealth of existing knowledge, thereby liberating them from manually searching the literature~\cite{vansanten2021microbial}. However, accessing these databases often necessitates expertise in specialized query languages, such as SQL, when navigating relational databases. Therefore, owing to the multidisciplinary nature of the field, natural product researchers are required to possess a robust background involving organic chemistry, microbiology, genetics, and, occasionally, computer science.

Recent advancements in large language models (LLMs) have transformed various scientific domains, including medical science~\cite{moor2023foundation}, chemistry~\cite{zheng2023chatgpt}, and geoscience~\cite{deng2024proceedings}. There has been a growing call for the application of AI technologies to natural product research, with a particular focus on building knowledge graphs~\cite{meijer2024empowering}. While LLMs are pretrained on extensive text corpora and possess foundational knowledge about natural products~\cite{openai2023gpt4}, their understanding remains limited. This is due to biochemical pathway complexity, diverse data types, sparse documentation, and data retrieval challenges. Additionally, without specialized biological tools, LLMs often produce hallucinated outputs that lack biological validity~\cite{xiao2024cellagent}. Fine-tuning can help LLMs adapt to specific tasks or domains by enhancing their grasp of specialized vocabulary and nuances~\cite{parthasarathy2024ultimate}, but challenges remain, such as the need for real-time knowledge updates, labelled data, computational resources, and self-assessment strategies. Furthermore, the absence of strategies for self-assessing output quality and the black-box nature of these approaches often lead to a lack of trust among users. This is especially critical in natural product research.

LLM-driven agents have shown significant potential for efficiently automating practical tasks~\cite{xiao2024cellagent, boiko2023emergent, hong2023metagpt}. Given their inherent autonomy, reactivity, proactiveness, and social capabilities, LLMs are ideally suited to function as the principal cognitive architectures or controllers of AI agents and to expand their perceptual and action spaces through strategies such as multimodal perception and tool utilization~\cite{xi2023rise}. However, when a single agent is tasked with handling complex processes, it often faces challenges because of the excessive number of steps required in a single run. This not only increases the instability of the execution process but also highlights the absence of robust mechanisms for monitoring output results, thus increasing the likelihood of producing uncontrolled or erroneous outputs10. Furthermore, because LLM-driven agents rely on text as their primary input‒output medium, they face limitations when addressing specialized fields, such as natural product research. In such domains, the complexity of information, ranging from biosynthetic pathways to structural and spectral data, makes it difficult to accurately convey professional terms and concepts through text alone.

\begin{figure}[t]
  \centering
  \includegraphics[width=\linewidth]{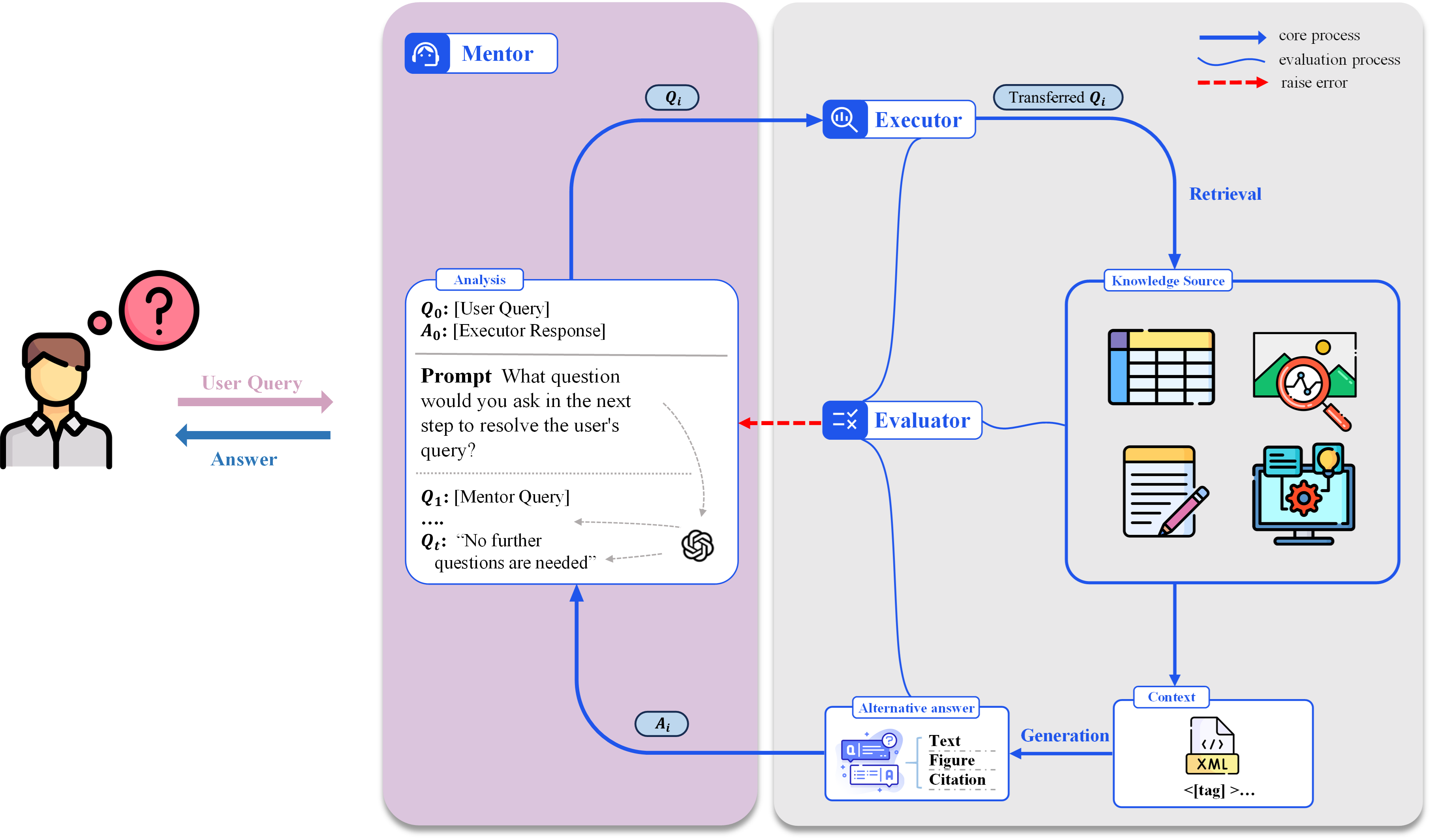}
    \setlength{\abovecaptionskip}{1pt}
    \caption{\scriptsize The workflow of ChatT2 framework. \textbf{Core process:} The Mentor and Executor engage in t rounds of interaction. In each round, the Mentor submits a query to the Executor, who processes it and retrieves relevant contextual information from various knowledge bases. This information may include text, images, citations, and other supporting materials to address the current query. Based on the user’s original intent and the results returned by the Executor in previous rounds, the Mentor formulates the next query. After completing t rounds, the Mentor summarizes all the questions and answers from the interaction into a final response, which is then returned to the user. \textbf{Evaluation process:} The Evaluator interacts with the Executor before, during, and after the retrieval and answer generation process to enhance response quality. \textbf{Raise error:} If the Evaluator identifies anomalies during the Executor’s retrieval process, it reports them to the Mentor.}
  \label{fig:workflow}
\end{figure}

Inspired by these advancements, we propose an LLM-driven multiagent framework for exploring natural products, incorporating generation-augmented methods to better align our results with scientific expectations. As a case study, we introduce ChatT2, a bacterial aromatic polyketide assistant that comprehends natural language tasks and completes complex queries through conversation. ChatT2 comprises three agents, namely, mentor, executor and evaluator each of which has defined responsibilities (Figure~\ref{fig:workflow}). To assess the performance of the multiagent framework, we developed NPBENCH, a benchmark spanning five key research scenarios involving the science of natural products. Using type II polyketides as an example, we designed 50 open-ended questions to construct NPBENCH (type II polyketide version) for the evaluation of ChatT2. To minimize manual involvement, we utilized both experts and LLMs to conduct a reasonable assessment. ChatT2 demonstrated significant improvements over the other tested models in terms of performance and occasionally showed remarkable response diversity, thus highlighting its potential as an intelligent assistant for natural product research. Our contributions in this work can be summarized as follows: (1) the creation of a construction schema of an external database that extends the capabilities of LLMs to the fields of biology and chemistry, thereby enhancing data processing efficiency and scientific accuracy; (2) the introduction of a multiagent framework designed to facilitate interdisciplinary integration between biology and chemistry, thus enabling collaboration across these scientific domains, with a particular focus on applications such as natural product research; and (3) the establishment of a robust and transferable evaluation standard to assess the performance of ChatT2, which was developed by our multiagent framework, thereby validating the framework itself.

\vspace{-3mm}

\section{Results}
\subsection{Overview of ChatT2}
\vspace{-2mm}

To address the challenges posed by the limited capacity of general LLMs to comprehend specialized corpora and manage complex natural product information, we developed a multiagent framework incorporating three distinct LLM-driven expert roles: mentor, executor, and evaluator. The mentor functions as an intermediary, devising plans, collaborating iteratively with the executor to gather and synthesize information, and delivering comprehensive outputs to the user. The executor processes inputs obtained from the mentor, leverages historical memory, customized databases and scientific tools and makes informed decisions to fulfill the requests of the mentor. The evaluator collaborates with the executor and mentor to prevent errors and hallucination information, thus ensuring the reliability of the output results (Figure~\ref{fig:workflow}). The pseudocode of this workflow is shown in Table S1. Using this framework, ChatT2 provides accurate, reliable, and traceable responses to user queries. Additionally, at an advanced application level, ChatT2 is employed for topic mining in natural product research, thus making it particularly suitable for tasks such as literature reviews and experimental assistance.

\textbf{\emph{Mentor:}} To solve a complex problem, a user may need to engage in multiple interactions with the LLM, pointing out misunderstandings in its responses or highlighting inadequacies in its temporary answers until the LLM provides the desired response. This requires the user to stay online and is not a user-friendly approach, particularly for beginners in specialized fields. To address this issue, ChatT2 introduces the mentor as an intermediary between the user and the executor of the system. Upon receiving an input from the user, the mentor interacts with the executor to perform a stepwise analytical process on the user’s query through the chain-of-thought (CoT) technique, which guides the executor to retrieve sufficient information, and then synthesizes a comprehensive response to present to the user (Figure~\ref{fig:workflow}). Additionally, the mentor plays a critical role in offering feedback to users about the anomalies identified by the evaluator, guiding them to refine their queries or elaborate on their problems (Figure~\ref{fig:workflow}). The functionality of the mentor mimics brain cognition, performing tasks such as thinking, discovery, and error correction. Its dynamic ability extends beyond simple query processing, enabling it to generate insights that transcend the original scope of the problem.

\textbf{\emph{Executor:}} To address the challenge of hallucinated information in domain-specific queries~\cite{huang2024survey}, the executor of ChatT2 employs a retrieval-augmented generation (RAG) framework to retrieve relevant information from external databases, which is then integrated with the contextual learning capabilities of LLMs to generate traceable responses~\cite{gao2024retrieval}. As shown in Figure~\ref{fig:workflow}, the executor is responsible for understanding the input acquired from the mentor and making action decisions on the basis of historical memory. When addressing the query of the mentor, the executor retrieves additional information by querying databases or invoking tools on the basis of the input of the mentor and historical conversations. The executor then combines the information derived from various sources into a cohesive text format (e.g., XML), which is used as the context input of the LLM to generate the final response. The final supporting context includes four parts: \emph{<Table>, <Text>, <Image>, <Tool>}.

\begin{figure}[!htbp]
  \centering
  \includegraphics[width=\linewidth]{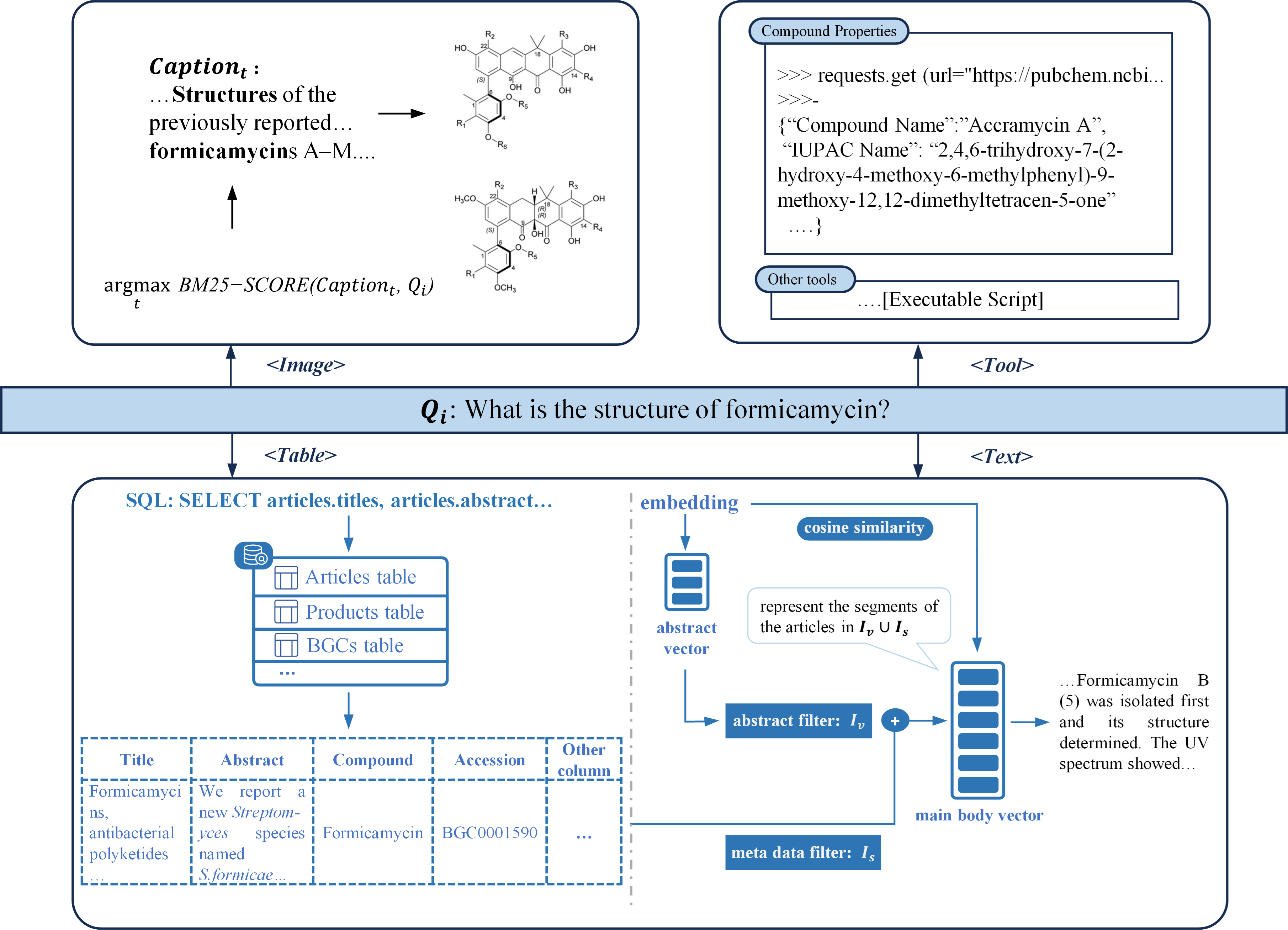}
    \setlength{\abovecaptionskip}{1pt}
    \caption{\scriptsize Multiple retrieval source of executor. When the Executor receives a query, it transforms the query into four different types of searches across a relational database, vector database, image database, and tool repository to retrieve relevant information. \textbf{Relational and Vector Databases:} The query is transformed into SQL statements and text embeddings: SQL statements are executed to retrieve table information related to the query, presented within \emph{<Table>} tags. During this process, relevant literature sources are gathered, forming a collection of papers denoted as $I_s$. Text embeddings are used to perform hierarchical semantic similarity matching based on cosine similarity. First, a semantically similar set of papers is identified, denoted as $I_v$. The embeddings of the query are then used to match the semantic similarity of document chunks from $I_v \cup I_s$, extracting text relevant to the query from the body of the papers. This information is presented within \emph{<Text>} tags. \textbf{Image Database:} The query is matched against image titles using BM25 scores. High-scoring images are identified, and their URLs are saved as image information, presented within \emph{<Figure>} tags. \textbf{Tool Repository:} The Executor parses key information from the query as input for prepackaged executable scripts. The execution results of these scripts are treated as tool repository information, encapsulated within \emph{<Tool>} tags.}
  \label{fig:executor}
\end{figure}

As shown in Figure~\ref{fig:executor}, during the process of retrieving information from the relational and vector databases, the query is transformed into SQL statements and text embeddings. The SQL statements are executed to retrieve table information related to the query, which is presented within \emph{<Table>} tags. Additionally, during this process, relevant literature sources are gathered, forming a collection of papers denoted as \emph{$I_s$}. The text embeddings are used to perform hierarchical semantic similarity matching on the basis of cosine similarity. First, a semantically similar set of papers is identified and denoted as $I_v$. The embeddings of the query are then used to match the semantic similarity of the document chunks acquired from $I_v \cup I_s$, extracting query-relevant text from the bodies of the papers. This information is presented within <Text> tags. For image database retrieval purposes, the query is matched against image titles via BM25 scores. High-scoring images are identified, and their URLs are saved as image information and presented within \emph{<Image>} tags. During tool repository retrieval, the executor parses the key query information as an input for prepackaged executable scripts. The execution results of these scripts are treated as tool repository information and encapsulated within \emph{<Tool>} tags. These four distinct databases ensure that the executor can access the maximum amount of traceable supporting material to respond effectively to the queries of the mentor. The detailed database construction has been implemented in the methods section.

\textbf{\emph{Evaluator:}} While various measures have been implemented in the executor to ensure that ChatT2 provides accurate answers to domain-specific questions, another critical focus is ensuring that ChatT2 can refuse to answer queries beyond its capabilities, thereby enhancing the credibility of its responses. This is especially important for knowledge-intensive tasks, such as question-answering applications. To address this, an additional agent, the evaluator, was developed. The evaluator interacts with the executor to perform three key tasks: refusing answers, filtering irrelevant information and tracking citations. These tasks occur in three stages: (1) preprocessing, which occurs when the user query is submitted; (2) mid-processing, which occurs during the retrieval of contextual information by the executor; and (3) postprocessing, which occurs during the generation of the final context by the executor. These three stages ensure the credibility of the final results and prevent the inclusion of erroneous or hallucinated information. For queries that ChatT2 cannot answer or for cases wherein the executor fails to retrieve relevant information, the evaluator provides feedback by refusing to answer. For queries that are answerable, the evaluator ensures that the executor includes correct citation sources in its final response, thereby enhancing the factual reliability and practical utility of the answers. This feature is particularly beneficial for researchers when analysing the results.

\vspace{-3mm}

\vspace{-4mm}
\subsection{Assessment of ChatT2}
\vspace{-2mm}

To evaluate the capabilities of ChatT2, we constructed a benchmark dataset named NPBENCH (type II polyketide version). This dataset consists of 50 open-ended queries that were designed to align with the needs of natural product researchers, particularly those focused on the main domains such as biosynthesis, structures, functions, applications, and classification (Table S2). The evaluation of ChatT2 was conducted through two tasks. (1) Response quality assessment. The quality of the responses generated by ChatT2 was compared with that of mainstream LLMs on the NPBENCH dataset without CoT prompting. In this case, ChatT2 performed only single-round retrieval, thus enabling the evaluation of its direct response capabilities. (2) In-depth analysis and exploration assessment. This task evaluated ChatT2's ability to perform problem decomposition and exploratory question answering via CoT prompts, and multiturn retrieval was performed using the NPBENCH dataset.

\begin{figure}[t]
  \centering
  \includegraphics[width=\linewidth]{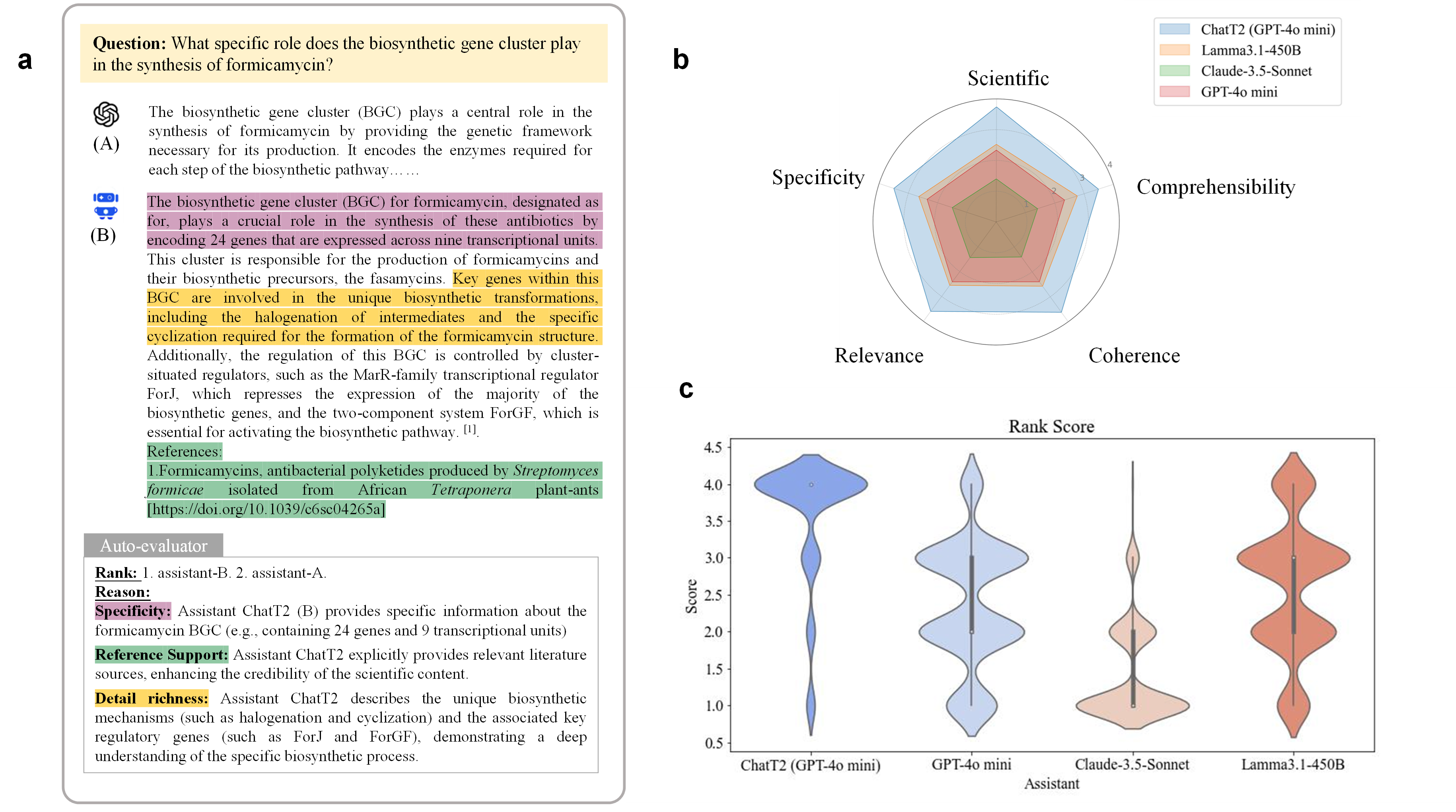}
    \setlength{\abovecaptionskip}{1pt}
    \caption{\scriptsize Comparison results of responses from different models and ChatT2. (a) The figure shows the comparison of the response texts between the original LLM and ChatT2, which uses the same LLM as its brain, for the same question; (b) Radar chart comparing the performance of responses from ChatT2 and different models across five evaluation metrics. Violin plot illustrating the rank score distributions of responses from ChatT2 and different models.}
  \label{fig:assessment}
\end{figure}

\textbf{\emph{ChatT2 provides more scientific and accurate answers:}} To evaluate the performance of ChatT2 and reduce its reliance on manual evaluations, we employed GPT-4 as an automated judge to evaluate the output responses, and we ranked the responses from most to least effective according to the imposed query requirements. We also provided explanatory rationales for these rankings based on five metrics: scientific accuracy, coherence, relevance, specificity, and comprehensibility. As shown in Figure~\ref{fig:assessment}a, for the same question, ChatT2 provided more comprehensive responses (with information granularity that was precise to the gene function level), cited the work from related articles, and included proper citations. By indexing additional databases, ChatT2 delivered more scientifically rigorous and instructive responses to natural product-related queries. After all the questions in NPBENCH were evaluated across various metrics, as shown in Figure~\ref{fig:assessment}b, ChatT2 (developed from GPT-4o mini) achieved higher rank scores than the baseline GPT-4o mini model. Furthermore, ChatT2 surpassed the other comparative models, demonstrating the efficacy of agent collaboration for enhancing the quality of LLM responses for type II polyketide-related inquiries. A detailed analysis revealed that the supporting materials retrieved were not only more relevant but also closely aligned with the preferences of both the LLM and human evaluations. Moreover, while the overall performance of GPT-4o mini on the current popular benchmarks was comparable to that of a 70-billion-parameter model, ChatT2 demonstrated superior performance in certain metrics compared with even a 405-billion-parameter model. These findings underscore the practicality and versatility of the ChatT2 framework in terms of optimizing smaller-scale models, making it well suited for diverse research environments in academia or laboratories. Additionally, we selected the winning rate and Spearman correlation coefficients to evaluate the automated judgements of the LLMs. Briefly, the winning rate is defined as the percentage of subtype questions for which the model ranks first. The Spearman correlation coefficient measures the consistency of the rank assignments produced by an LLM when the same question is evaluated repeatedly (r times). For each question subtype, a positive Spearman correlation coefficient indicates consistent ranking results for the corresponding LLM. Notably, the Spearman correlation coefficients for repeated evaluations (r times) across different question types all exceeded 0.6, suggesting that the rankings provided by the tested LLM as an automatic judge were reliable and trustworthy. As shown in Table~\ref{tab:winning_rate_spearman}, ChatT2 demonstrated a significant advantage across all categories in the NPBENCH dataset, outperforming GPT-4o mini by more than four times. Notably, ChatT2 achieved substantial improvements for the biosynthesis, function and application categories, which are of particular interest to NP researchers.

\begin{table}[htbp]
\centering
\caption{\scriptsize Winning rate of different models and spearman correlation across different types of questions.}
\label{tab:winning_rate_spearman}
\begin{tabular}{lccccc}
\toprule
\multirow{2}{*}{Assistant} & \multicolumn{5}{c}{Question type} \\
\cmidrule(lr){2-6}
 & Biosynthesis & Structure & Function & Application & Classification \\
\midrule
ChatT2 (GPT-4o mini) & \textbf{0.844} & \textbf{0.712} & \textbf{0.808} & \textbf{0.804} & \textbf{0.624} \\
GPT-4o mini & 0.044 & 0.052 & 0.052 & 0.072 & 0.164 \\
Claude-3.5-Sonnet & 0 & 0.016 & 0 & 0.008 & 0 \\
Llama 3.1-405B & 0.112 & 0.22 & 0.14 & 0.116 & 0.212 \\
\midrule
Mean spearman correlation\textsuperscript{(r=5)} & 0.714 & 0.623 & 0.621 & 0.647 & 0.607 \\
\bottomrule
\end{tabular}
\end{table}

\vspace{-3mm}
\begin{figure}[t]
  \centering
  \includegraphics[width=\linewidth]{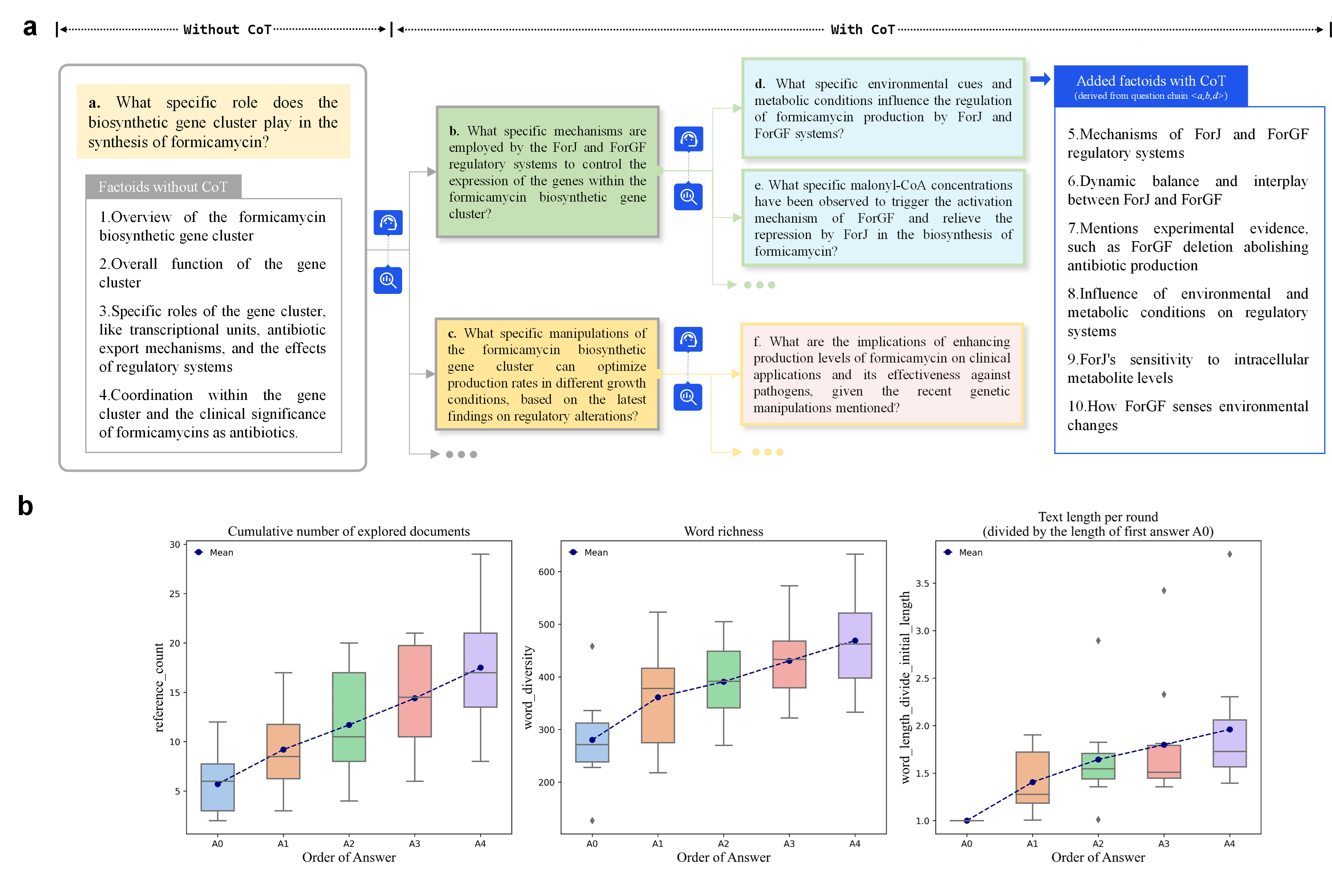}
    \setlength{\abovecaptionskip}{1pt}
    \caption{\scriptsize The overall result of ChatT2 with CoT and without CoT between the Mentor and Executor. (a) The panel shows the questions formulated by the Mentor over multiple interaction rounds with the Executor; (b) After applying CoT techniques, the evaluation results for the answers are presented in terms of the cumulative number of explored documents, word richness, and text length per round.}
  \label{fig:Mentor and Executor}
\end{figure}

\textbf{\emph{ChatT2 has potential for exploring interesting research topics:}} The responses generated by the mentor and executor through multiple iterations implemented via CoT techniques are rich, which is attributed to the problem decomposition and exploratory question-answering capabilities of ChatT2. The richness of the responses produced by ChatT2 encompasses two primary aspects: question diversity and answer richness. Question diversity refers to the number of semantic question clusters generated during the iterative process, with each cluster representing a distinct exploration branch from the mentor-generated questions for a given initial query. Questions that are phrased differently but address the same topic are grouped within the same semantic cluster. As illustrated in Figure~\ref{fig:Mentor and Executor}a, during the iterative process, the mentor dynamically generates new questions on the basis of the results provided by the executor in the previous round and the CoT prompts. Despite the use of identical initial questions, the subsequent question-generation trajectories of the mentor and executor exhibit a significant degree of divergence. For example, the focus may shift from delving into gene mechanisms (green to blue) to exploring optimization strategies for producing natural products and their clinical applications (yellow to pink).This underscores the potential for multiturn retrieval to expand the breadth of inquiries while enabling deeper explorations of user intents. Furthermore, as the number of iterations increases, the number of distinct question paraphrase clusters increases, resulting in more diverse responses. The collaborative strategy between the mentor and executor effectively broadens the sampling space of the text generation process in the generated language model, significantly enhancing the richness and informativeness of the generated content.

Answer richness evaluates the quality of the responses produced by ChatT2 via metrics such as the number of referenced documents, the degree of word richness, and the text length per round. An increase in the number of referenced documents during the iterative process indicates meaningful exploratory progress, whereas a lack of growth suggests that ChatT2 may have reached a bottleneck or convergence state. Additionally, word richness, defined by the number of unique words used, complements text length by providing insights into the depth and detail of ChatT2's answers. To validate the capabilities of multiturn retrieval assisted by the CoT technique, we utilized ChatT2 to conduct five iterative rounds of testing on subset of type II polyketide-related questions derived from NPBENCH, and we analysed the cumulative number of explored documents, word richness, and text length at each round. As shown in Figure~\ref{fig:Mentor and Executor}b, when a single iteration was performed, the average number of documents referenced by ChatT2 was fewer than five. However, as the number of iterations increased, the number of referenced documents increased correspondingly. Similarly, the variety of words and the length of the generated text increased with the number of iterations. By the end of the iterative process, the responses of ChatT2 were significantly more comprehensive and enriched with a greater number of referenced sources than the outputs derived from a single iteration.

\begin{figure}[t]
  \centering
  \includegraphics[width=\linewidth]{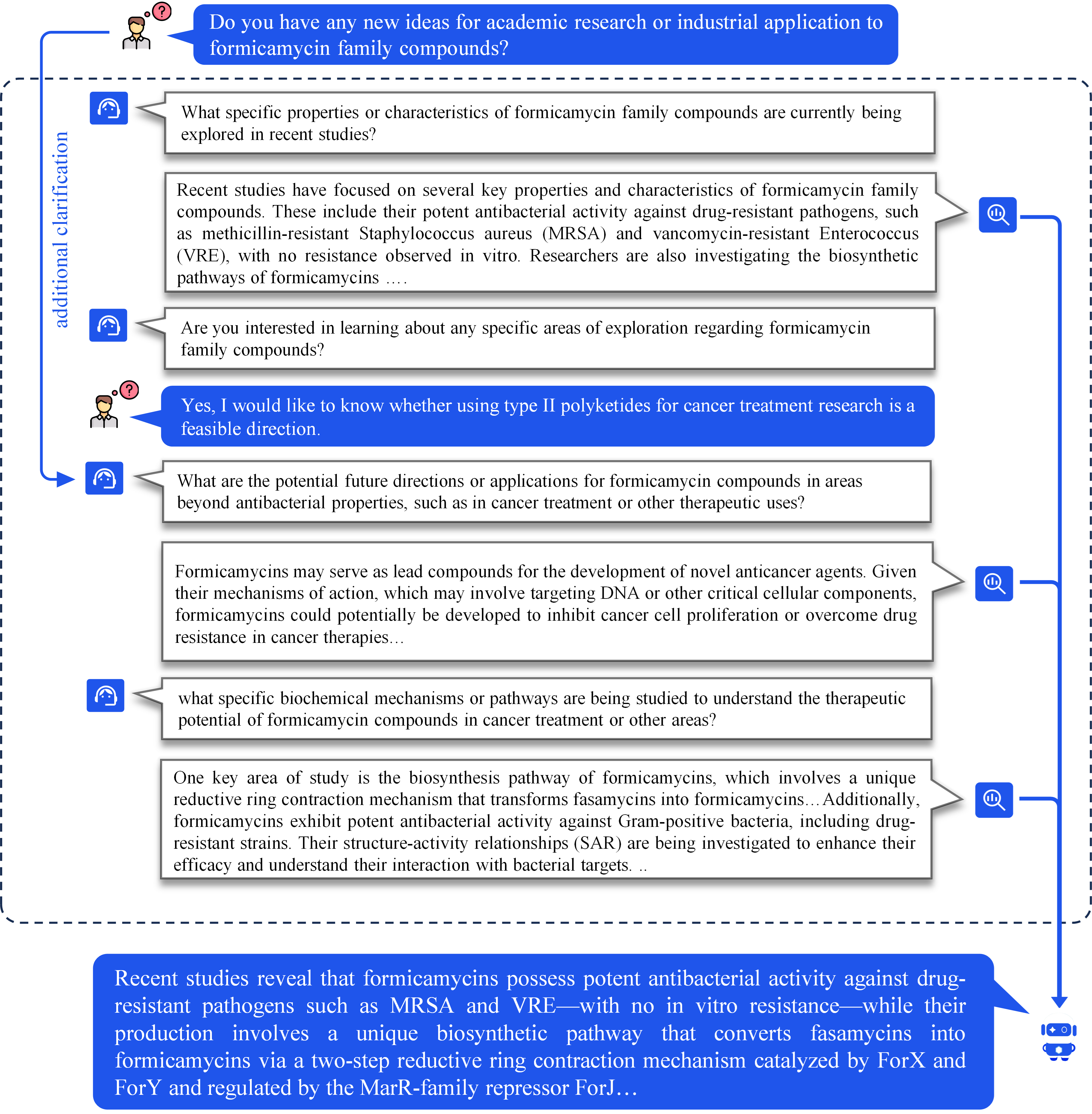}
    \setlength{\abovecaptionskip}{1pt}
    \caption{\scriptsize An open-ended question-answer demonstration operating in Auto mode.}
  \label{fig:Auto mode}
\end{figure}

Effectively utilizing CoT techniques for multiturn retrieval poses significant challenges for mentors. To enhance the user experience, we propose three modes for deployment: Fixed, Auto, and Updated. In Fixed mode, a static reasoning chain is constructed on the basis of only the initial query; in Auto mode, the mentor proactively assesses user preferences and optimizes the chain of thought using the preferences provided by the user during the interaction; and in Updated mode, the chain is reconstructed during each interaction to reflect the most current requirements. Detailed information is provided in the Methods section. To demonstrate the ability of ChatT2 to serve as a research assistant, we initiated an investigation by posing an open-ended question in Auto mode: “Do you have any new ideas for academic research or industrial application to formicamycin family compounds?” As illustrated in Figure~\ref{fig:Auto mode}, upon receiving this question, ChatT2 activated its multiturn task scheme and, with the assistance of CoT techniques, proceeded to deconstruct the query. Initially, guided by the mentor, ChatT2 formulated the first subquestion: “What specific properties or characteristics of formicamycin family compounds are currently being explored in recent studies?” The executor and evaluator subsequently retrieved and processed pertinent information to address this subquestion, mirroring the human approach of establishing a foundational understanding of a domain prior to engaging with broader issues. After sufficient background information on the formicamycin family was acquired, ChatT2 asked the user, “Are you interested in learning about any specific areas of exploration regarding formicamycin family compounds?” This step is analogous to that of human interlocutors seeking clarification to refine the focus of the discussion. Integrating the retrieved information with the user’s feedback, ChatT2 then proposed two additional questions: (1) “What are the potential future directions or applications for formicamycin compounds in areas beyond antibacterial properties, such as in cancer treatment or other therapeutic uses?” and (2) “What specific biochemical mechanisms or pathways are being studied to understand the therapeutic potential of formicamycin compounds in cancer treatment or other areas?” These subsequent inquiries were designed to accrue further data necessary for addressing the initial query. Through this iterative, multiturn retrieval process, ChatT2 provided deeper insight for expanding natural products for academic research or industrial applications. When faced with open-ended questions, ChatT2 first deconstructed the question, collected background information, revisited the user’s specific requirements, and then integrated and refined the knowledge through successive rounds of dialogue. Through this process, ChatT2 exhibited a certain human-like inference style, reflecting a trend towards anthropomorphism in its conversational behaviour. Overall, our results demonstrated the ability of ChatT2 to generate high-quality, traceable, and scientifically relevant responses, thus highlighting its potential for accelerating knowledge dissemination and supporting chemists in developing personalized research assistants.

\vspace{-3mm}

\section{Discussion}

In this study, we presented an innovative LLM-driven multiagent framework tailored for exploring natural products, emphasizing the discovery and identification of bacterial aromatic polyketides. ChatT2, which was developed within a multiagent framework, is an intelligent assistant designed to comprehend natural language tasks and address complex queries through interactive conversations. In contrast to the previously published agent framework, ChatT2 retains only three essential agent roles and leverages the CoT along with a “question-generation–response–question” approach to ensure that each action step is well founded rather than solely relying on the task decomposition capabilities of LLMs. Furthermore, ChatT2 not only accepts user tasks during the initial stages but also proactively solicits assistance from users during task execution, such as providing results from practical experiments or seeking further clarification of objectives. To assess the efficacy of our multiagent framework, we developed NPBENCH, a comprehensive benchmark encompassing five critical research scenarios within the field of natural product research. The results demonstrated that ChatT2 not only outperformed the existing models in terms of accuracy and reliability but also remarkably met the expectations of natural product scientists, thus underscoring its potential as a powerful intelligent assistant for researchers. The successful implementation and evaluation of ChatT2 highlight the transformative potential of integrating multiagent systems with advanced language models to facilitate and accelerate scientific discoveries.

Several limitations remain in the development of ChatT2 within the multiagent framework. First, from a data perspective, the external knowledge database schema still requires significant manual effort to filter and expand content for various natural product subfields. From a model perspective, ChatT2 heavily relies on its underlying LLM for intelligence. Second, although ChatT2 is highly effective in type II polyketide-related tasks, its understanding of other domains needs further validation. Third, as the model’s internal knowledge base expands, reliance on external knowledge retrieval diminishes. Consequently, there is potential to shift the focus of the framework towards individual reasoning and collaborative discussion, which could further enhance response efficiency.

Future research could focus on expanding the external database to include a more diverse and comprehensive range of natural products, thereby broadening the applicability of the framework. Additionally, integrating specialized agents to address more complex and domain-specific tasks would enhance the versatility and problem-solving capabilities of the system. Beyond natural product research, this framework could be adapted and applied to other fields within biotechnology and chemistry, paving the way for interdisciplinary innovations. Finally, improving the ability of the framework to incorporate real-time data and dynamically adapt to evolving scientific paradigms will be essential for maintaining its relevance and usability. Such advancements will ensure that the platform remains a cutting-edge tool in the rapidly changing landscape consisting of natural product research and related scientific domains.

\section{Experimental Section}
\subsection{Design of the multiagent framework}
\vspace{-2mm}

ChatT2 was developed within an autonomous multiagent framework comprising a mentor, an executor, and an evaluator to generate expert information about natural products. The mentor functions as an intermediary, devising plans, collaborating iteratively with the executor to gather and synthesize information, and delivering comprehensive outputs to the user. The executor processes the inputs obtained from the mentor; leverages historical memory, vector databases or tools; and makes informed decisions to fulfil the requests of the mentor. The evaluator ensures the reliability of the results by preventing errors and hallucinated information. GPT-4o mini was selected as the brain of the agents because of its 128K context window, faster inference speed, developer-friendly API, and JSON output capability, which enables stable interagent communication~\cite{openai2024gpt4omini}. The overall preferences and configuration of ChatT2 are shown in Table S3 and the prompt content corresponding to the specific functions of each agent is provided in Figure S1 and Table S4.

\vspace{-4mm}
\subsection{Description of the Mentor}
\vspace{-2mm}

LLMs can conduct in-context few-shot learning through CoT prompting~\cite{brown2020language, schick2021exploiting, wei2022chain}. Thus, we implemented a CoT prompt in the mentor agent to process five research scenarios: biosynthesis, chemical modification and structural optimization, functions, applications, and classification. The input of the mentor includes the following: \emph{<prompt, initial question, subsequent conversation, research approach (CoT)>}. For each query, the mentor processes these inputs to generate sequential follow-up questions and produces a comprehensive summary report.

\vspace{-4mm}
\subsection{Description of the Executor}
\vspace{-2mm}

The executor employs a multi-database architecture comprising relational databases, vector databases, image databases, and tool repositories for information retrieval. This component operates under three key principles: (1) information diversity optimization, (2) fault-tolerant retrieval mechanisms, and (3) context signal-to-noise ratio maximization. These design principles ensure efficient and reliable data access across different query types.

\textbf{\emph{Relational Database:}} Relational Database: We constructed a relational database by integrating data from MIBiG and PubChem, supplemented with literature-mined and expert-curated information. The database schema consists of four primary tables (articles, products, BGCs, and genes) and two junction tables (article descriptions and product biosynthesis). The schema implements many-to-many relationships between article-product and BGC-product tables, enabling comprehensive querying of type II polyketide natural products, their biosynthesis-related gene clusters, and the associated literature. The detailed structure of the relational database is presented in Figure S2.

\textbf{\emph{Vector Database:}} We developed a vector dataset via literature from the Web of Science and MIBiG databases, with a focus on type II polyketide products. The text processing pipeline begins by extracting plain text from PDFs via the pdfminer Python package, followed by the removal of fragmented characters, watermarks, and non-content elements. Figure captions were filtered via regular expressions for separate image database construction. Additionally, we employed a hybrid chunking method that integrated fixed-length chunking and overlap chunking across different segmentations. The text chunks were then converted into embeddings for calculating the semantic similarity of queries via the cosine distance metric in the retrieval-augmented generation process. The chunk size was aligned with the maximum input length of the embedding model. The same model was used to obtain embeddings of user queries in the ChatT2 workflow to ensure consistency of semantic representations in the vector space.

A hierarchical vector database was constructed to optimize the retrieval of relevant text sequences. The database architecture consisted of two layers: article abstracts and their corresponding text chunks. Text embeddings were generated for both abstracts and chunks, with each chunk containing both text content and citation information. As shown in Figure~\ref{fig:executor}, the retrieval process involved a two-step approach: matching query embeddings with article abstracts, and searching within the chunks of matched articles. Supplementary methods 1.2.1 shows the detailed implementation of the vector database and JSON schema of this vector database is illustrated in Figure S3.

\textbf{\emph{Image Database:}} Two complementary image databases were developed to enhance visual outputs: a manually curated static database and a dynamically generated database through tool invocation. For image retrieval, BM25 algorithm was implemented to match user queries with image captions, as traditional semantic embedding methods are less suitable for keyword-based caption text. Each database entry contained both image captions and URLs, with relevant URLs being included in the LLM's response when corresponding images matched user queries. Supplementary methods 1.2.2 shows the detailed implementation of the image database.

\textbf{\emph{Tool Repository:}} External tools were integrated into the system through LLM function calling capabilities. ChatT2 incorporates multiple sequence alignment tools for processing nucleotide or amino acid sequences in FASTA format, employing sophisticated global alignment algorithms (ClustalW, MUSCLE, MAFFT) to identify conserved regions, insertions or deletions, and sequence similarity patterns. Complementing this functionality, a gene function annotation tool analyzes user-provided sequences against type II polyketide repositories, utilizing bit-score and identity metrics to identify functionally conserved regions and active sites for comprehensive biological characterization. Additionally, the PubChem integration component enables chemical information retrieval with secondary filtering mechanisms to ensure result relevance. Tool selection was guided by prompt-based instructions, enabling the LLM to determine appropriate tool usage on the basis of user queries (Table S5). The outputs from these tools were subsequently incorporated into the retrieval context for final response generation.

\textbf{\emph{Robustness of Fault-Tolerance Retrieval:}} ChatT2 allows users to pose questions in various languages, despite the underlying English-based information repository. The relational database implementation included fuzzy search functionality that accommodated insertion‒deletion errors within specified ranges and effectively handled common misspellings of compound names, article names, and gene names to maintain high recall rates. To address potentially vague or imprecise queries, particularly from novice users, the mentor will ask users for providing extra demand. This approach integrated dialogue history with the retrieved context and enabled the system to request clarification when the user intent was unclear. This dynamic question-refinement strategy enhances the quality of user‒system interactions and leads to more accurate and relevant responses.

\textbf{\emph{Maximizing the Signal-to-Noise Ratio of the Context:}} To optimize RAG performance, we focused on maximizing the signal-to-noise ratio in the final generation phase. This was achieved by providing the LLM with as much relevant information as possible while avoiding the inclusion of irrelevant data. This approach is critical for reducing the occurrence of hallucinated responses, which typically emerge when the system processes excessive irrelevant information~\cite{yu2024proceedings}.

\textbf{\emph{Filtering the Retrieved Text:}} We implemented a similarity-based filtering mechanism to reduce the irrelevant context provided to the LLM. The embeddings of the text chunks were prestored in a vector database. During retrieval, chunks with semantic similarity values below a threshold MIN\_SCORE\_THRESHOLD (Table S3) were filtered out before extracting the document content to <Text> tags.

\textbf{\emph{User-Defined Recall Rate:}} We developed a dynamic recall rate mechanism to accommodate varying information retrieval requirements. Instead of implementing a fixed number of chunks that will be selected in final context, we introduce a self-controlled recall rate (SCRR) that users can adjust according to their needs. The value of SCRR is related to the length of the relevant context obtained through retrieval. Users can opt for a low SCRR for domain-specific questions or a high SCRR for general questions.

\vspace{-4mm}
\subsection{Description of the Evaluator}
\vspace{-2mm}

The ChatT2 evaluator plays a critical role in ensuring the quality, relevance, and accuracy of the responses generated by the system. It performs three primary functions: refusing answers, filtering irrelevant information, and tracking citations. These functions are essential for maintaining the integrity and reliability of the information provided to users.

\textbf{\emph{Refusing to Answer:}} Cheng et al. reported that supervised fine-tuning (SFT) can lead to the constructed model being overly conservative, resulting in incorrect refusals to answer known questions~\cite{cheng2024can}. To address this problem, we implemented an evaluator for query assessment. Rather than relying on the executor or mentor components, the evaluator determined ChatT2's ability to address user queries on the basis of prompts.

\textbf{\emph{Irrelevant Information Detection:}} We design a context refinement pipeline where the executor submits user queries and their associated structured context to the evaluator postretrieval. The evaluator, guided by prompts, filtered out XML elements with empty or irrelevant content, ensuring that the retained content of XML maintained relevance to the query. When all the elements were deemed irrelevant and removed, the system triggered a feedback mechanism to the executor. Upon receiving this feedback, the executor either initiated a new retrieval process or prompted the user for additional information or query reformulation.

\textbf{\emph{Citation Tracking:}} This task was designed to evaluate whether the answers generated by the executor, which included citations, had the correct reference sources. In the final generated answer, each supporting chunk could fall into one of four categories: (1) a chunk was used, and its content was cited; (2) a chunk was used, but its content was not cited; (3) a chunk was not used, yet its content was cited; and (4) a chunk was not used, and its content was not cited. In this part, the evaluator was primarily responsible for assessing the results acquired from the executor for the presence of situations 2 and 3. If either situation existed, the evaluator provided a guidance prompt to the executor to regenerate its answer with citations. The executor then followed this guidance to produce a revised answer.

\vspace{-4mm}
\subsection{Benchmark dataset}
\vspace{-2mm}

To evaluate the performance of ChatT2 in natural product research scenarios, we developed NPBENCH, a specialized benchmark dataset focused on type II polyketides. Unlike traditional generated language model benchmarks that assess general language understanding and generation capabilities~\cite{wang2024mmlu, hendrycks2020measuring}, NPBENCH specifically evaluates practical problem-solving abilities in natural product research contexts. We structured NPBENCH around five key research categories: biosynthesis, chemical modification and structural optimization, functions, applications, and classification. We designed 50 representative problems across these categories, with examples detailed in Table S2. Each problem was crafted to have definitive expected answers, thus enabling a clear assessment of model knowledge boundaries in the natural products domain. The evaluation framework incorporates both single-turn and multiturn assessments, moving beyond traditional benchmark metrics23 to focus on practical research value and potential for novel insights. We supplemented automated evaluation to assess response quality and relevance to real-world research scenarios.

\vspace{-4mm}
\subsection{Assessment Tasks}
\vspace{-2mm}

\textbf{\emph{Single-turn assessment task without CoT:}} We implemented a hybrid evaluation framework combining LLM-based assessment and expert validation to address the challenges of comparing natural language outputs in the natural products domain. For single-turn assessments, we employed GPT-4 as the primary evaluator, tasking it with ranking responses from ChatT2 and various LLMs across five metrics: scientific accuracy, coherence, relevance, specificity, and comprehensibility. To increase the credibility of the results, GPT-4 provides detailed explanations for its rankings. The evaluation process utilized a standardized prompt template incorporating question text, anonymized model outputs (labelled “assistants A-D”), and evaluation metrics. Additionally, human experts can sample and review the ranking results of the GPT-4 to ensure its validity. We implemented a scoring system where assistants received points on the basis of their rank position: the best-ranked assistant received a score equal to the number of assistants, whereas the lowest-ranked assistant received a score of 1.

Zheng et al. demonstrated that when GPT-4 is used for ranking, the results are susceptible to positional bias~\cite{li2024proceedings}. Pairwise comparisons with multiple order swaps can lead to a quadratic increase in workload with the number of elements to be ranked~\cite{zheng2023judging}. To mitigate the influence of positional factors on the ranking results, we set the temperature to 0 and instructed GPT-4 to perform repeated rankings five times (r = 5), with the order of answers randomly shuffled before each repetition. We also introduced the Spearman correlation coefficient as a consistency metric for repeated rankings.

The Spearman correlation coefficient, ranging from -1 to 1, served as a reliability indicator for the ranking process. A high correlation suggests that the GPT-4 provides consistent evaluations of shuffled answers, demonstrating independence from positional factors. Conversely, a low Spearman correlation indicates that GPT-4 struggles to correctly rank the answers, in which case human experts take over the ranking evaluation process.

\textbf{\emph{Multi-turn assessment task with CoT:}} For the same question, the responses generated by the mentor and executor through multiple iterations are often difficult to reproduce. Therefore, we developed a two-dimensional evaluation framework for multiturn assessments that focuses on question diversity and answer richness to analyse the iterative response generation capabilities of ChatT2.

Question diversity was measured through semantic clustering analysis, where expert reviewers categorized mentor-generated questions into distinct exploration branches. Questions addressing the same topic, despite different phrasing, were grouped into single semantic clusters. Answer richness was evaluated via three key metrics: the number of referenced documents, word richness, and text length per round. Referenced documents reflect the breadth of knowledge that is objectively incorporated into the self-reflection and iterative dialogue processes of ChatT2. Vector databases serve as essential components of the final context, as not all questions require external resources such as image databases or tool repositories. An increasing number of referenced documents across iterations indicates valuable exploratory progress. We implemented two termination conditions for the iterative process: a convergence state, which was detected when the number of referenced documents plateaus; and mentor assessment, which was triggered when sufficient information is gathered to address the initial query comprehensively. Word richness measures the total number of unique words used until the current iteration. In addition to text length, word richness provides a complementary perspective on the depth and detail of the answers yielded by ChatT2.

\vspace{-4mm}
\subsection{Multiagent Historical Memory Management}
\vspace{-2mm}

We designed a hierarchical message lifecycle management system to optimize the dialogue flow between different agents in ChatT2. The system maintains separate historical message contexts on the basis of specific agent interactions and their requirements. As shown in Figure~\ref{fig:workflow}, we established three distinct lifecycle levels to manage message history. The highest-level lifecycle governs user-mentor interactions, beginning when users initiate conversations with ChatT2. The secondary lifecycle manages mentor-executor dialogues, starting when the mentor forwards user queries or generates new queries to the executor and ending upon the mentor's confirmation of satisfactory query resolution. The lowest-level lifecycle controls evaluator’s communication, which are specifically designed for certain evaluation task completion. This hierarchical structure ensures that each agent accesses only the relevant historical context. By maintaining independent message streams between different entities, the system prevents information overlap and ensures efficient task completion for each agent.
\section*{Acknowledgements}

This work was supported by the National Natural Science Foundation of China (32170079 to Z.Q., 32200035 to H.Z., and 32400235 to J.H.), the Natural Science Foundation of Guangdong (2024A1515012593 to Z.Q. and 2023A1515110175 to J.H.), Guangdong Talent Scheme (2021QN020100 to Z.Q.), Digital-Empowered Education Reform Project of Beijing Normal University at Zhuhai (jx2025108 to H.Z.).
\section*{Conflict of Interest}

The authors declare no competing financial interests.
\section*{Author Contributions}

Zhiwei Qin, Jiaquan Huang and Heqian Zhang designed and supervised the research. Yihan Wang, Qiandi Gao and Liangjun Ge performed the bioinformatic and established the algorithm. Yihui Zhuang designed the user interface. All authors analysed and discussed the data. Zhiwei Qin, Jiaquan Huang and Yihan Wang wrote the manuscript and all authors edited.
\section*{Data Availability Statement}

The data, materials and code supporting the findings reported in this study are available at GitHub repository: \url{https://github.com/Qinlab502/chatT2}. We also developed a Web site that allows users from any computer background to use it, which can be accessed at \url{https://chatt2.site/#/chat}.

\bibliographystyle{splncs04}
\bibliography{mybibliography}

\end{document}